\def\year{2021}\relax

\documentclass[letterpaper]{article} 
\usepackage{aaai21}  
\usepackage{times}  
\usepackage{helvet} 
\usepackage{courier}  
\usepackage[hyphens]{url}  
\usepackage{graphicx} 
\usepackage{natbib}  
\usepackage{caption} 
\usepackage[switch]{lineno}

\usepackage{multirow}

\usepackage[export]{adjustbox}
\usepackage{amsmath}
\usepackage{amssymb}
\usepackage{bm}
\usepackage[acronym]{glossaries}
    \newacronym{gan}{GAN}{Generative Adversarial Network}
    \newacronym{mscoco}{MS-COCO}{Microsoft Common Objects in Context}
    \newacronym{vg}{VG}{Visual Genome}
    \newacronym{clstm}{ConvLSTM}{\textit{convolutional} Long Short-Term Memory}
\usepackage{empheq}
\usepackage{enumitem}
\usepackage{tabularx}

\usepackage{xr}
\makeatletter
\newcommand*{\addFileDependency}[1]{
  \typeout{(#1)}
  \@addtofilelist{#1}
  \IfFileExists{#1}{}{\typeout{No file #1.}}
}
\makeatother

\newcommand*{\myexternaldocument}[1]{%
    \externaldocument{#1}%
    \addFileDependency{#1.tex}%
    \addFileDependency{#1.aux}%
}
\myexternaldocument{supplementary}

\title{Object-Centric Image Generation from Layouts}
\author{Tristan Sylvain,\textsuperscript{\rm 1,2} Pengchuan Zhang,\textsuperscript{\rm 3} Yoshua Bengio,\textsuperscript{\rm 1,2,4} R Devon Hjelm,\textsuperscript{\rm 3,1} Shikhar Sharma\textsuperscript{\rm 5}}  
\affiliations{
    \textsuperscript{\rm 1}Mila, Montr\'eal, Canada \qquad
    \textsuperscript{\rm 2}Universit\'e de Montr\'eal, Montr\'eal, Canada \qquad
    \textsuperscript{\rm 3}Microsoft Research\\
    \textsuperscript{\rm 4}CIFAR Senior Fellow \qquad
    \textsuperscript{\rm 5}Microsoft Turing\\ 
    \{tristan.sylvain, yoshua.bengio\}@mila.quebec, \{penzhan, devon.hjelm, shikhar.sharma\}@microsoft.com 
}

\begin{document}

\maketitle

\begin{abstract}
We begin with the hypothesis that a model must be able to understand individual objects and relationships between objects in order to generate complex scenes with multiple objects well. Our layout-to-image-generation method, which we call Object-Centric Generative Adversarial Network (or OC-GAN), relies on a novel Scene-Graph Similarity Module (SGSM). The SGSM learns representations of the spatial relationships between objects in the scene, which lead to our model's improved layout-fidelity. We also propose changes to the conditioning mechanism of the generator that enhance its object instance-awareness. Apart from improving image quality, our contributions mitigate two failure modes in previous approaches: (1) spurious objects being generated without corresponding bounding boxes in the layout, and (2) overlapping bounding boxes in the layout leading to merged objects in images. Extensive quantitative evaluation and ablation studies demonstrate the impact of our contributions, with our model outperforming previous state-of-the-art approaches on both the COCO-Stuff and Visual Genome datasets. Finally, we address an important limitation of evaluation metrics used in previous works by introducing SceneFID -- an object-centric adaptation of the popular Fr{\'e}chet Inception Distance metric, that is better suited for multi-object images.
\end{abstract}


\section{Introduction}
\glspl{gan}~\cite{gan} have been at the helm of significant recent advances in image generation~\cite{gan,dcgan,wgan-gp,cgan-proj,biggan}. 
Apart from unsupervised image generation, \gls{gan}-based image generation approaches have done well at conditional image generation from labels~\cite{dcgan,sagan,biggan}, captions~\cite{reed-icml,stackgan,attngan,objgan,sd-gan}, conversations~\cite{chatpainter,geneva,storygan}, scene graphs~\cite{sg2im,interactive-sg2im,soarisg}, layouts~\cite{layout2im,isrls}, segmentation masks~\cite{spade}, \emph{etc}. 
While the success in single-domain or single-object image generation has been remarkable, generating complex scenes with multiple objects is still challenging.

\begin{figure}[ht]
    \centering
    \newcommand{\figwidth}{0.195\columnwidth}
    \newcommand{\figtwidth}{0.19\columnwidth}
    
    \begin{center}
        \hfill \parbox{\figtwidth}{\centering Layout} \hfill \parbox{\figtwidth}{\centering SPADE} \hfill \parbox{\figtwidth}{\centering SOARISG} \hfill \parbox{\figtwidth}{\centering LostGAN} \hfill \parbox{\figtwidth}{\centering OC-GAN (ours)} \hfill\\
    
        \includegraphics[width=\figwidth]{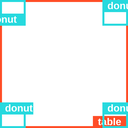}\hfill
        \includegraphics[width=\figwidth]{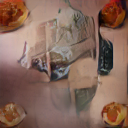}\hfill
        \includegraphics[width=\figwidth]{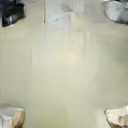}\hfill
        \includegraphics[width=\figwidth]{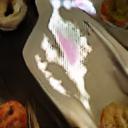}\hfill
        \includegraphics[width=\figwidth]{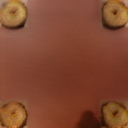}\\
        
        \includegraphics[width=\figwidth]{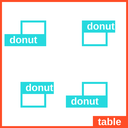}\hfill
        \includegraphics[width=\figwidth]{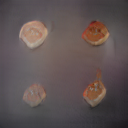}\hfill
        \includegraphics[width=\figwidth]{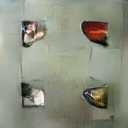}\hfill
        \includegraphics[width=\figwidth]{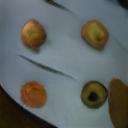}\hfill
        \includegraphics[width=\figwidth]{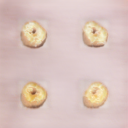}\\
        
        \includegraphics[width=\figwidth]{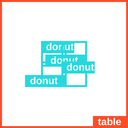}\hfill
        \includegraphics[width=\figwidth]{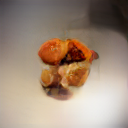}\hfill
        \includegraphics[width=\figwidth]{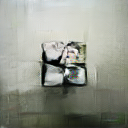}\hfill
        \includegraphics[width=\figwidth]{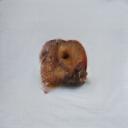}\hfill
        \includegraphics[width=\figwidth]{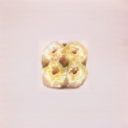}
    \end{center}
    \vspace*{-\normalbaselineskip}
    \caption{Each row depicts a layout and the corresponding images generated by various models. Along each column, the donuts converge to the centre. In addition to more clearly defined objects, our method is the only one that maintains distinct objects for the final layout, for which bounding boxes slightly overlap.}
    \label{fig:coco-converging-donuts}
\end{figure}

\begin{figure}[!ht]
    \centering
    \newcommand{\figwidth}{0.245\columnwidth}
    \newcommand{\figtwidth}{0.24\columnwidth}
    \begin{center}
        \hfill \parbox{\figtwidth}{\centering Layout} \hfill \parbox{\figtwidth}{\centering SOARISG} \hfill \parbox{\figtwidth}{\centering LostGAN} \hfill \parbox{\figtwidth}{\centering Ours} \hfill \\
        
        \includegraphics[width=\figwidth]{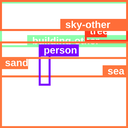}\hfill
        \includegraphics[width=\figwidth]{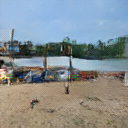}\hfill
        \includegraphics[width=\figwidth]{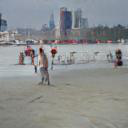}\hfill
        \includegraphics[width=\figwidth]{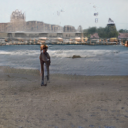}\hfill \\
        \includegraphics[width=\figwidth]{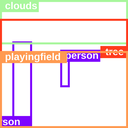}\hfill
        \includegraphics[width=\figwidth]{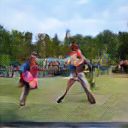}\hfill
        \includegraphics[width=\figwidth]{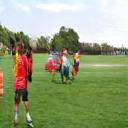}\hfill
        \includegraphics[width=\figwidth]{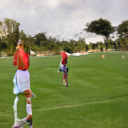}
    \end{center}
    \vspace*{-\normalbaselineskip}
    \caption{Existing models introduce spurious objects not specified in the layout, a failure mode over which our model improves significantly.}
    \label{fig:coco-128-spurious}
\end{figure}

Generating realistic multi-object scenes is a difficult task because they have many constituent objects~\citep[e.g., the Visual Genome dataset,][can contain as many as 30 different objects in an image]{krishna2017visual}.
Past methods focus on different input types, including scene graphs~\cite{sg2im,soarisg}, pixel-level semantic segmentation~\cite{objgan}, and bounding box-level segmentation~\cite{layout2im,isrls}. In addition, some methods also consider multi-modal data, such as instance segmentation alongside pixel-wise semantic segmentation masks~\cite{spade,hd-pix2pix}.
Orthogonal to input-related considerations, methods tend to rely on additional components to help with the complexity of scene generation, such as attention mechanisms~\cite{attngan,objgan} and explicit disentanglement of objects from the background~\cite{finegan}.

Despite these advances, models still struggle in creating realistic scenes. As shown in Figs.~\ref{fig:coco-converging-donuts} and~\ref{fig:coco-128-spurious}, even simple layouts can result in merged objects, spurious objects, and images that do not match the given layout (low layout-fidelity). 
To counter this, we propose Object-Centric GAN (OC-GAN), an architecture to generate realistic images with \emph{high layout-fidelity} and \emph{sharp objects}.
Our primary contributions are:
\begin{itemize}[nolistsep]
    \item We introduce a set of novel components that are well-motivated and improve performance for complex scene generation. Our proposed scene-graph-based retrieval module (SGSM) improves layout-fidelity. We also introduce other improvements, such as conditioning on instance boundaries, that help generating sharp objects and realistic scenes.
    \item Our model improves significantly on the previous state of the art in terms of a set of classical metrics. In addition to standard metrics, we also perform a detailed ablation study to highlight the effect of each component, and a human evaluation study to further validate our findings.
    \item We discuss the validity of the metrics currently used to evaluate layout-to-image methods, and building on our findings, motivate the use of SceneFID, a new evaluation setting which is more adapted to multi-object datasets.
\end{itemize}

\section{Related Work}
\subsubsection{Conditional scene generation}
For some time, the image generation community has focused on scenes that contain multiple objects in the foreground~\cite{reed-icml,stackgan,sg2im}. Such scenes, which can contain large amount of objects of very different scales, are very complex relative to single-object images. Several conditional image generation tasks have been formulated using different subsets of annotations. Text-based image generation using captions~\cite{reed-icml,stackgan,attngan,objgan,sd-gan} or even multi-turn conversations~\cite{chatpainter,geneva,storygan} have gained significant interest. However, with increasing numbers of objects and their relationships in the image, understanding long textual captions becomes difficult~\cite{sg2im,chatpainter}. 
Text-based image generation approaches are also not immune to small perturbations in text leading to quite different images~\cite{sd-gan}.

\subsubsection{Layout-based synthesis}
Generating images from a given layout makes the analysis more interpretable by decoupling the language understanding problem from the image generation task. Another advantage of generating from layouts is more controllable generation: it is easy to design interfaces to manipulate layouts. In this work we will focus on coarse layouts, where the scene to be generated is specified by bounding-box-level annotations.
Layout-based approaches fall into 2 broad categories. Some methods take scene-graphs as inputs, and learn to generate layouts as intermediate representations~\cite{sg2im,soarisg}. In parallel, other approaches have focused on generating directly from coarse layouts~\cite{isrls,layout2im}. Models that perform well on fine-grained pixel-level semantic maps also can be easily applied to this setting~\cite{spade,pix2pix,hd-pix2pix}.
Almost all recent approaches have in common the use of \emph{patch} and \emph{object discriminators} (to ensure whole image and object quality). In addition to this, image quality has been improved by the addition of \emph{perceptual losses}~\cite{spade,soarisg,hd-pix2pix}, \emph{multi-scale patch-discriminators}~\cite{spade}, which motivate some of our architecture choices. Finally, modulating the parameters of batch- or instance-normalization layers~\cite{ioffe2015batch,ulyanov2016instance} with a function of the input condition can provide significant gains, and this is done per-channel in~\cite{odena2017conditional} or per pixel~\cite{spade,isrls}. As bounding box layouts are coarse for this task, it is common to introduce \emph{unsupervised mask generators}~\cite{isrls,ma2018exemplar} to provide estimated shapes for this conditioning.

Finally, there is a growing body of literature involving semi-parametric~\cite{qi2018semi,li2019pastegan} models that use ground-truth training images to aid generation. We consider the case of such models in the Appendix.

\subsubsection{Scene-graphs and image matching}
Scene graphs are an object-centric representation that can provide an additional useful learning signal when dealing with complex scenes.
Scene-graphs are often used as intermediate representations in image captioning~\cite{yang2019auto,anderson2016spice}, reconstruction~\cite{gu2019scene} and retrieval~\cite{johnson2015image}, as well as in sentence to scene graph~\cite{schuster2015generating} and image to scene graph prediction~\cite{lu2016visual,newell2017pixels}.

 By virtue of being a simpler abstraction of the scene than a layout, they emphasize \emph{instance awareness} more than layouts which focus on pixel-level class labels. Secondly, for scenarios that might require generating multiple diverse images, they provide more variability in reconstruction and matching tasks as the mapping from a scene graph to an image is one to many usually. These points explain their use in higher-level visual reasoning tasks such as visual question answering~\cite{teney2017graph} and zero-shot learning~\cite{sylvain2020locality, sylvain2020zero}, and also motivate the use of scene graph-based retrieval in our model. In our work, we generate scene graphs depicting positional relationships (such as ``to the left of'', ``above'', ``inside'', \emph{etc.}) from given spatial layouts and leverage them to learn the relationships between objects, which would be more difficult for a model to distill from pixel-level layouts. 

There has been strong interest in image and caption similarity modules for retrieval~\cite{fang2015captions,huang2013learning} and for text-to-image generation, most recently with the DAMSM model proposed in~\cite{attngan}. Despite similar interest in scene graph to image retrieval~\cite{johnson2015image,quinn2018semantic}, and the large improvements in text-to-image synthesis resulting from the DAMSM~\cite{attngan,objgan}, our approach is the first to use a scene graph to image retrieval module when training a generative model.

\section{Proposed Method}
\subsection{Scene-Graph Similarity Module}

\begin{figure}
    \centering
    \includegraphics[width=0.9\columnwidth]{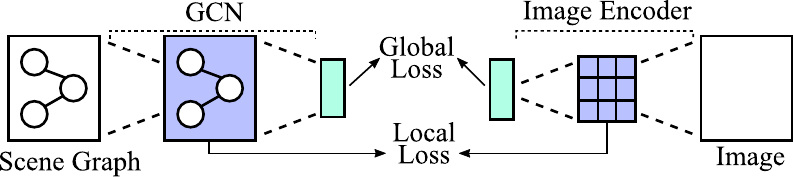}
    \caption{The SGSM module. The SGSM module computes similarity between the scene-graph and the generated image, providing fine-grained matching-based supervision between the positional scene-graph and the generated image.}
    \label{fig:sgsm}
\end{figure}

We introduce the Scene Graph Similarity Module (SGSM) as a means of increasing the \emph{layout-fidelity} of our generated images.
This multi-modal module, described summarily in Fig.~\ref{fig:sgsm}, takes as input an image and a scene-graph (nodes corresponding to objects, and edges corresponding to spatial relations). We extract \emph{local visual features}  $\bm{v}_{i}$ from the \emph{mixed\_6e} layer in an Inception-V3 network~\cite{szegedy2016rethinking} pre-trained on the ImageNet dataset. We extract \emph{global visual features} $\bm{v}^G$ from the final pooling layer. We encode the graph using a Graph Convolutional Network~\citep[GCN,][]{goller1996learning} to obtain \emph{local graph features} $\bm{g}_{j}$ and apply a set of graph convolutions followed by a graph pooling operation to obtain \emph{global graph features} $\bm{g}^G$. Note that each local and global feature is extracted and linearly projected to a common semantic space. In what follows, $\cos$ is the cosine similarity, and the $\gamma_k$s are normalization constants. We use $L/G$ when the local and global terms are interchangeable.
We use the modified dot-product attention mechanism of~\citet{attngan} to compute the \emph{visually attended local graph embeddings} $\tilde{\bm{g}}_j$:
\begin{align}
    \bm{s}_{ij} &= \gamma_1 \frac{\exp \big({\bm{g}_j}^T \bm{v}_i\big)}{\sum_{i'}\exp \big({{\bm{g}_j}^T \bm{v}_{i'}}\big)}, &
    \tilde{\bm{g}}_j &= \frac{ \sum_{i} \exp(\bm{s}_{ij}) \bm{v}_i }{\sum_{i} \exp(\bm{s}_{ij})}
\end{align}

Then we can define a \emph{local similarity metric} between the source graph embedding ${\bm{g}}_j$ and the visually aware local embedding ${\tilde{\bm{g}}_j}$ similar to~\citet{attngan}. Intuitively, the similarity will be strong when the source graph embedding is close to the visually aware embedding. This local similarity will encourage different patches of the image to match the objects expected from the scene graph. The \emph{global similarity metric} is classically the cosine distance between embeddings:
\begin{empheq}[left=\empheqlbrace]{align}
   \text{Sim}^L(S, I') &= \log \Big(\sum_j \exp\big(\gamma_2 \cdot \cos({\tilde{\bm{g}}_j}, {\bm{g}}_j) \big)\Big)^{\frac{1}{\gamma_2}}\\
   \text{Sim}^G(S, I') &= \cos\big(\bm{v}^G,      {\bm{g}^G}\big)
\end{empheq}

Finally we can define a global and local probability model in a similar way to e.g.~\citet{huang2013learning}:
\begin{align}
    \mathbb{P}^{L/G}(S, I') \propto \exp\Big(\gamma_3 \cdot \text{Sim}^{L/G} (S, I')\Big)
\end{align}

Normalizing over the images or scenes in the batch $B$ (negative examples are selected by mis-matching the image and scene-graph pairs in the batch) leads to e.g.: $\mathbb{P}^{L/G}(S | I) = \frac{ \mathbb{P}^{L/G}(S, I)}{\sum_{I' \in B} \mathbb{P}^{L/G}(S, I')}$. We define the loss  terms as the log posterior probability of matching an image I and \emph{the corresponding} scene graph (and vice-versa):

\begin{empheq}[left=\empheqlbrace]{align}
   \mathcal{L}_{L/G} &= -\log \mathbb{P}_{L/G}(S | I) - \log \mathbb{P}_{L/G}(I | S) \\
   \mathcal{L}_{\text{SGSM}} &= \mathcal{L}_{L} + \mathcal{L}_{G}
\end{empheq}

Empirically, the SGSM resulted in large gains in performance as shown in Table~\ref{table:ablation}. Our hypothesis is that the scene graph, in a similar way to a caption, provides easier, simpler to distil relational information contained in the layout, which results in stronger performance compared to generation using just the layout.
Architectural details of the SGSM and related data processing are described in the Appendix.

\subsection{Instance-Aware Conditioning}
\label{subsec:instance-boundary}
\begin{figure}[h]
    \centering
    \includegraphics[width=0.45\textwidth]{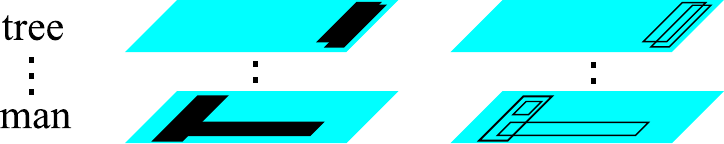}
    \caption{Blue indicates 0 and black indicates 1. (Left) The per-class mask constructed from the layout by many previous methods makes it impossible to distinguish unique object instances in several cases. (Right) Our mask consists of instance boundaries making it easier for the model to distinguish unique object instances using no extra information than already contained in the layout.}
    \label{fig:instance-aware}
\end{figure}
As in~\citet{spade,isrls}, the parameters $\gamma, \beta$ of our batch-normalization layers are \emph{conditional} and determined on a per-pixel level \citep[as opposed to classical conditional batch-normalization,][]{de2017modulating}. In our case, these parameters are determined by three concatenated inputs: \emph{masked object embeddings, bounding-box layouts and bounding-box instance boundaries}. Masked object embeddings~\cite{ma2018exemplar,isrls} and bounding-box layouts (using 1-hot embeddings) have been previously used in the layout to image setting. A shortcoming of these conditioning inputs is that they do not provide any way to distinguish between objects of the same class if their bounding boxes overlap. We use the layout's bounding-box boundaries, shown in Figure~\ref{fig:instance-aware}, as additional conditioning information. The addition of the bounding-box instance boundaries helps the model in mapping overlapping conditioning semantic masks to separate object instances, the absence of which led previous state-of-the-art methods to generate merged outputs as shown in the donut example in Fig.~\ref{fig:coco-converging-donuts}. Importantly, the instance boundaries do not add any additional information compared to the baselines: (1) they are bounding-box rather than fine-grained boundaries, and (2) instance information is already available to other models (Layout2Im and LostGAN have object-specific codes as an example). Rather, adding these boundaries acts like a prior encouraging our model to focus on generating distinct objects.

\subsection{Architecture}
Our OC-GAN model is based on the \gls{gan} framework. The generator module generates the images conditioned on the ground-truth layout. The discriminator predicts whether the input image is generated or real. The discriminator has an additional component which has to discriminate objects present in the input image patches corresponding to the ground-truth layout object bounding boxes. We present an overview of the model in Fig.~\ref{fig:model-overview} and describe the components below. Additional details are in the Appendix.

\begin{figure*}
    \centering
    \includegraphics[width=0.9\textwidth]{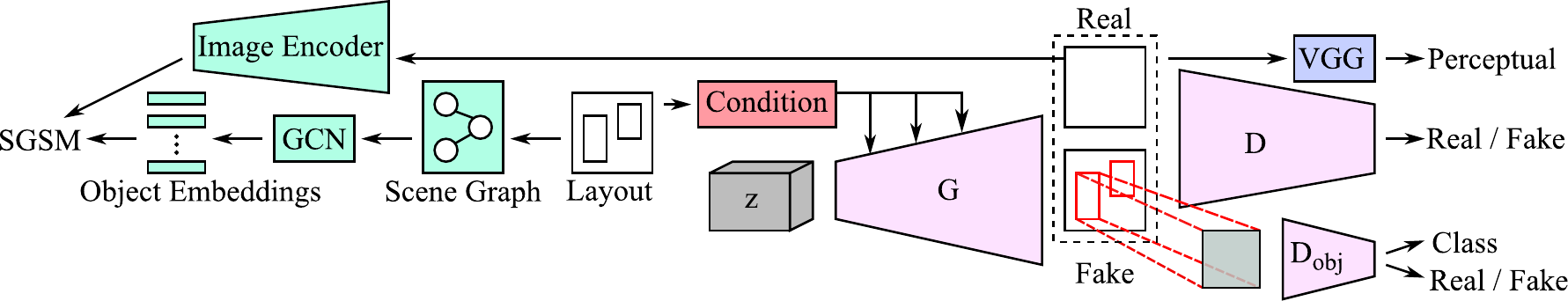}
    \caption{Overview of our OC-GAN model. The GCN and Image Encoder modules are trained separately and then frozen. The condition for the Generator's normalization and the Scene Graph encoding the spatial relationships between objects are both derived from the input layout. The SGSM and the instance-aware normalization lead our model to generate images with higher layout-fidelity and sharper, distinct objects. The `Condition' box corresponds to the three inputs listed in the subsection on the instance-aware conditioning.}
    \label{fig:model-overview}
\end{figure*}

\subsubsection{Generator}
As a means of disentangling our model's performance from a specific choice of generator architecture, we used a classical residual~\cite{he2016deep} architecture consisting of 4 layers for $64\times64$ inputs, and 5 layers for $128\times128$ inputs, as used recently in~\citet{spade,isrls,hd-pix2pix}. The residual decoder $G$ takes as input image-level noise. As described in the previous section, we further condition the generation by making the normalization parameters of the batch-norm layers of the decoder dependent on the layout and instance boundaries.

\subsubsection{Discriminator}
We use two different types of discriminators, an object discriminator, and a set of patch-wise discriminators. The object discriminator $D_{obj}$ takes as input crops of the objects (as identified by their input bounding boxes) in real and fake images resized to size $32\times32$ and is trained using the Auxiliary-Classifier~\citep[AC,][]{odena2017conditional} framework, resulting in a classification and an adversarial loss. Next, two \emph{patch-wise discriminators} $D^p _1, D^p _2$ output estimates of whether a given patch is consistent with the input layout. We apply them to the original image and the same image down-sampled by a factor of 2 (no weight sharing) in a similar fashion to \citet{spade,hd-pix2pix}.\\

\subsection{Loss Functions}
In the following, $x$ denotes a real image, $l$ a layout, and $z$ noise. We also denote objects with $o$ and their labels $y_o$.

\paragraph{Perceptual loss}
Adding a perceptual loss~\cite{dosovitskiy2016generating,gatys2016image,johnson2016perceptual} to our model improved results slightly. We extract features using a VGG19 network~\cite{simonyan2014very}. The loss has expression: $\mathcal{L}_P = \mathbb{E}_{x, l, z}\sum_{i=1}^N \frac{1}{D_i} ||F^{(i)}(x) - F^{(i)}(G(l, z))||_1$ where $F^{(i)}$ extracts the output at the i-th layer of the VGG and $D_i$ is the dimension of the flattened output at the i-th layer. 

\paragraph{Generator and Discriminator losses}
We train the generator and patch discriminators using the adversarial hinge loss~\cite{lim2017geometric}:
\begin{align}
    \mathcal{L}_G ^{\text{GAN}} &= -\mathbb{E}_{l, z} \Big[D_1 ^p(G(l, z), l) + D_2 ^p(G(l, z), l) \Big]\\
\begin{split}
    \mathcal{L}_{D^p} &= \sum_{i=1} ^2 \Big\{-\mathbb{E}_{x, l} \Big[\min(0, -1 + D_i ^p(x, l)) \Big]\\
    &\qquad-\mathbb{E}_{l, z} \Big[\min(0, -1 - D_i ^p(G(l, z), l)\Big] \Big\}
\end{split}
\end{align}
The object discriminator follows the AC-GAN framework, leading to $\mathcal{L}_G ^{AC}$ and $\mathcal{L}_{D_{obj}} ^{AC}$. The final expression is:

\begin{align}
    \mathcal{L}_{G} &= \mathcal{L}^{\text{GAN}}_{G} + \lambda_P \mathcal{L}_\text{P} + \lambda_{\text{SGSM}}\mathcal{L}_\text{SGSM} + \lambda_{\text{AC}} \mathcal{L}_{G} ^{AC}\\
    \mathcal{L}_{D} &= \mathcal{L}_{D^p} +  \lambda_o \mathcal{L}_{D_{obj}} ^{AC}
\end{align}

We fix $\lambda_P=2, \lambda_o=1, \lambda_{\text{SGSM}}=1, \lambda_{\text{AC}}=1$ in our experiments.

\section{Experiments}
\subsection{Datasets}
We run experiments on the COCO-Stuff~\cite{caesar2016coco} and \gls{vg}~\cite{krishna2017visual} datasets which have been the popular choice for layout- and scene-to-image tasks as they provide diverse and high-quality annotations. The former is an expansion of the \gls{mscoco} dataset~\cite{mscoco}. 
We apply the same pre-processing and use the same splits as ~\citet{sg2im,layout2im}. The summary statistics of the two datasets are presented in the appendix, Table~\ref{tb:datasets}. 

Our OC-GAN model takes three different inputs:
\begin{itemize}
    \item The spatial layout \emph{i.e.} object bounding boxes and object class annotations.
    \item Instance boundary maps computed directly from the layout. While they appear redundant once the bounding boxes are provided, they aid the model in better differentiating different objects especially different instances of the same object class.
    \item Scene-graphs. These are constructed from the objects and spatial relations inferred from the bounding box positions following the setup in~\cite{sg2im}. While \gls{vg} provides more complex scene graphs, we restricted ourselves to spatial relations only for compatibility between the two datasets.
\end{itemize}

\subsection{Implementation and Training Details}
\label{sec:implementation}
Our code is written in PyTorch~\cite{paszke2017automatic}. We apply Spectral Normalization~\cite{miyato2018spectral} to all the layers in both the generator and discriminator networks. Each experiment ran on 4 V100 GPUs in parallel.
We use synchronized BatchNorm (all summary statistics are shared across GPUs). 

We used the Adam~\cite{kingma2014adam} solver, with $\beta_1$ = 0.5, $\beta_2$ = 0.999. The global learning rate for both generator and discriminators is 0.0001. $128\times128$ models and above were trained for up to 300\,000 iterations, $64 \times 64$ models were trained for up to 200\,000 iterations (early stopping on a validation set). The SGSM module is trained separately for 200 epochs. It is then fixed, and the rest of the model is trained.

\subsection{Baselines}
We consider all recent methods that allow layout-to-image generation (Layout2Im~\cite{layout2im}, LostGAN~\cite{isrls}, LostGAN-v2~\cite{lostganv2}). We report results for scene-graph-to-image methods (SG2Im~\cite{sg2im}, SOARISG~\cite{soarisg}) evaluated with \emph{ground-truth layouts} for a fair comparison. Finally, methods originally designed for generation from pixel-level semantic segmentation maps (SPADE~\cite{spade} and Pix2PixHD~\cite{hd-pix2pix}) are also considered as they can be readily adapted to this new context.



\subsection{Evaluation}
Evaluation of GANs is a complex issue, and the subject of a vast body of literature. In this paper, we focus on three existing evaluation metrics: Inception Score (IS)~\cite{salimans2016improved}, Fr{\'e}chet Inception Distance (FID)~\cite{heusel2017gans} and Classification Accuracy (CA). For the CA score, a ResNet-101~\cite{he2016deep} network is trained on object crops obtained from the real images of the train set of the corresponding dataset, as suggested by~\cite{soarisg}. The FID metric computes the 2-Wasserstein distance between the real and generated distributions, and therefore serves as an efficient proxy for the diversity and visual quality of the generated samples. While the FID metric focuses on the whole image, the CA metric allows us to demonstrate the ability of our model to generate realistic-looking objects within a scene. Finally, we include the Inception Score as a legacy metric.

\subsubsection{Our proposed metric: SceneFID}
We note that there exist many concerns in the literature regarding the use of metrics that are not designed or adapted to the task at hand. The Inception Score has been criticised~\cite{barratt2018note}, notably due to issues caused by the mismatch between the domain it was trained on (the ImageNet dataset comprising single objects of interest) and the domain of VG and COCO-Stuff images (comprising multiple objects in complex scenes), making it a potentially poor metric to evaluate generative ability of models in our setting. While the FID metric was introduced in response to Inception Score's criticisms, and was shown empirically to alleviate some of the concerns with it~\cite{im2018quantitatively,xu2018empirical,lucic2018gans}, it still suffers from problems in the layout-to-image setting. In particular, the single manifold assumption behind FID was found in~\citet{liu2018improved} to be problematic in a multi-class setting. This is \textit{a fortiori} the case in a multi-object setting as in VG and COCO. While~\cite{liu2018improved} introduce a class-aware version of FID, this is not applicable to our setting. We introduce the \emph{SceneFID} metric, where we compute the FID on the crops of all objects, resized to same size (224 $\times$ 224), instead of on the whole image. Thus, the SceneFID metric measures FID in the single manifold assumption it was designed for and extends it to the multi-object setting.

In addition to the above quantitative metrics, we also perform qualitative assessment of the model, notably by considering the effect of modifying the input layout on the output image.

\begin{figure}[!ht]
    \centering
    \newcommand{\figwidth}{0.18\columnwidth}
    \newcommand{\figtwidth}{0.18\columnwidth}
    \begin{center}
    \hspace{0.5em} 
        \hfill \parbox{\figtwidth}{\centering Layout} \hfill \parbox{\figtwidth}{\centering SPADE} \hfill \parbox{\figtwidth}{\centering SOARISG} \hfill \parbox{\figtwidth}{\centering LostGAN} \hfill \parbox{\figtwidth}{\centering OC-GAN} \hfill\\
        (a) \hfill
        \includegraphics[valign=m,width=\figwidth]{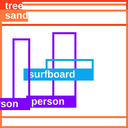}\hfill
        \includegraphics[valign=m,width=\figwidth]{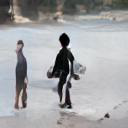}\hfill
        \includegraphics[valign=m,width=\figwidth]{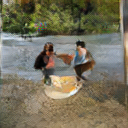}\hfill
        \includegraphics[valign=m,width=\figwidth]{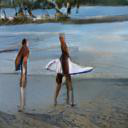}\hfill
        \includegraphics[valign=m,width=\figwidth]{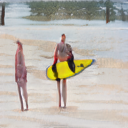}\\

        (b) \hfill        
        \includegraphics[valign=m,width=\figwidth]{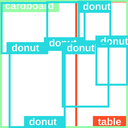}\hfill
        \includegraphics[valign=m,width=\figwidth]{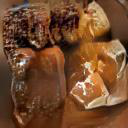}\hfill
        \includegraphics[valign=m,width=\figwidth]{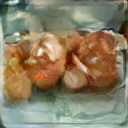}\hfill
        \includegraphics[valign=m,width=\figwidth]{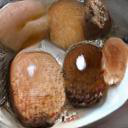}\hfill
        \includegraphics[valign=m,width=\figwidth]{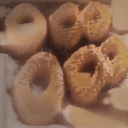}\\
        
        (c) \hfill
        \includegraphics[valign=m,width=\figwidth]{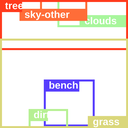}\hfill
        \includegraphics[valign=m,width=\figwidth]{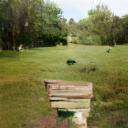}\hfill
        \includegraphics[valign=m,width=\figwidth]{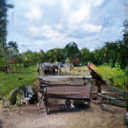}\hfill
        \includegraphics[valign=m,width=\figwidth]{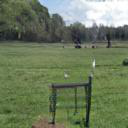}\hfill
        \includegraphics[valign=m,width=\figwidth]{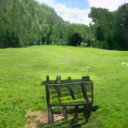}\\
        
        (d) \hfill
        \includegraphics[valign=m,width=\figwidth]{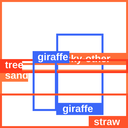}\hfill
        \includegraphics[valign=m,width=\figwidth]{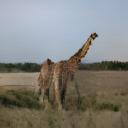}\hfill
        \includegraphics[valign=m,width=\figwidth]{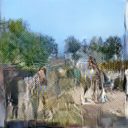}\hfill
        \includegraphics[valign=m,width=\figwidth]{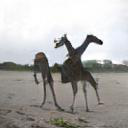}\hfill
        \includegraphics[valign=m,width=\figwidth]{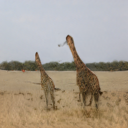}\\
        
        (e) \hfill
        \includegraphics[valign=m,width=\figwidth]{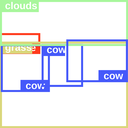}\hfill
        \includegraphics[valign=m,width=\figwidth]{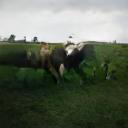}\hfill
        \includegraphics[valign=m,width=\figwidth]{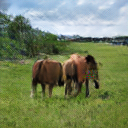}\hfill
        \includegraphics[valign=m,width=\figwidth]{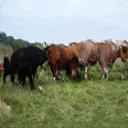}\hfill
        \includegraphics[valign=m,width=\figwidth]{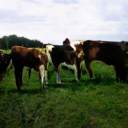}
    \end{center}
    \vspace*{-\normalbaselineskip}
    \caption{$128 \times 128$ COCO-Stuff test set images, taken from our method (OC-GAN), and multiple competitive baselines. Note the overall improved visual quality of our samples. In addition, for (d, e) many baselines introduce spurious objects, and for (b, d, e) spatially close objects are poorly defined and sometimes fused for the baselines.}
    \label{fig:coco-128-compare-models}
\end{figure}

\subsection{Quantitative Results}
We report comparisons of our model's performance to the set of all recent state-of-the-art methods. Where applicable and possible, we use metric values reported by the authors of the papers. SOARISG~\cite{soarisg} depends on semantic segmentation maps being available, and therefore it was not feasible to include results on VG for this method. Some papers introduced additional data-augmentation, such as LostGAN~\cite{isrls} which introduced flips of the real images during training. Where applicable, we report results using the same experimental setup as the authors, and highlight it in the results table. For all models that do not report CA scores, we evaluate them using images generated with the pre-trained models provided by their authors.

Table~\ref{table:results-table} shows that our model consistently outperforms the baselines in terms of IS, FID and CAS, often significantly. We note that for some models, the CAS score is above that reported for ground-truth images. This is due to the fact that a sufficiently capable generator will start to generate objects that are both realistic, and of the same distribution as the training distribution, rather than the test one.

\begin{table*}
\resizebox{\textwidth}{!}{%
\begin{tabular}{|l|l|c|c|c|c|c|c|}
\hline
\multicolumn{1}{|l|}{} &
  \multicolumn{1}{c|}{\multirow{2}{*}{Methods}} &
  \multicolumn{2}{c|}{Inception Score $\uparrow$} &
  \multicolumn{2}{c|}{FID $\downarrow$} &
  \multicolumn{2}{c|}{CA $\uparrow$} \\
 
\multicolumn{1}{|l|}{} &
  \multicolumn{1}{c|}{} &
  COCO &
  \multicolumn{1}{l|}{VG} &
  COCO &
  \multicolumn{1}{l|}{VG} &
  COCO &
  \multicolumn{1}{l|}{VG} \\ \hline
\multirow{3}{*}{Real Images}               & $64\times64$                                                 & 16.3 $\pm$ 0.4          & 13.9 $\pm$ 0.5            & 0                 & 0                 & 54.48            & 49.57            \\
                                           & $128 \times 128$                                             & 22.3 $\pm$ 0.5          & 20.5 $\pm$ 1.5            & 0                 & 0                 & 60.71            & 56.25            \\ 
 & $256 \times 256$  & 28.10 $\pm$ 0.5 & 28.6 $\pm$ 1.2& 0 & 0 & 63.04 & 60.40\\ \hline
\multirow{6}{*}{$64\times64$}              & SG2Im~\cite{sg2im}$\dagger$            & 7.3 $\pm$ 0.1           & 6.3 $\pm$ 0.2             & 67.96             & 74.61             & 30.04            & 40.29            \\ 
                                           & Pix2PixHD~\cite{hd-pix2pix}            & 7.2 $\pm$ 0.2           & 6.6 $\pm$ 0.3             & 59.95             & 47.71             & 20.82            & 16.98            \\ 
                                           & SPADE~\cite{spade}                     & 8.5 $\pm$ 0.3           & 7.3 $\pm$ 0.1             & 43.31             & 35.74             & 31.61            & 23.81            \\ 
                                           & Layout2Im~\cite{layout2im}$\dagger$    & 9.1 $\pm$ 0.1           & 8.1 $\pm$ 0.1             & 38.14             & 31.25             & 50.84            & 48.09            \\ 
                                           & SOARISG~\cite{soarisg}$\ast$ $\dagger$ & 10.3 $\pm$ 0.1          & N/A                       & 48.7              & N/A               & 46.1             & N/A              \\ 
                                           & OC-GAN (ours)                                                & $\mathbf{10.5 \pm 0.3}$          & $\mathbf{8.9 \pm 0.3}$             & $\mathbf{33.1}$              & $\mathbf{22.61}$             & $\mathbf{56.88}$            & $\mathbf{57.73}$            \\ \hline
\multirow{1}{*}{$64\times64$}              & LostGAN~\cite{isrls} (flips) $\dagger$ & 9.8 $\pm$ 0.2           & 8.7 $\pm$ 0.4             & 34.31             & 34.75             & 37.15            & 27.1             \\ 
\multirow{1}{*}{with flips}                & OC-GAN (ours)                                                & $\mathbf{10.8 \pm 0.5}$ & $\mathbf{9.3 \pm 0.2}$    & $\mathbf{29.57}$  & $\mathbf{20.27}$  & $\mathbf{60.39}$ & $\mathbf{60.79}$ \\ \hline
\multirow{5}{*}{$128\times128$}            & Pix2PixHD~\cite{hd-pix2pix}            & 10.4 $\pm$ 0.3          & 9.8 $\pm$ 0.3             & 62                & 46.55             & 26.67            & 25.03            \\ 
                                           & SPADE~\cite{spade}                     & 13.1 $\pm$ 0.5          & 11.3 $\pm$ 0.4            & 40.04             & 33.29             & 41.74            & 34.11             \\ 
                                           & Layout2Im~\cite{layout2im} $\diamond$  & 12.0 $\pm$ 0.4          & 10.1 $\pm$ 0.3            & 43.21             & 38.21             & 49.06            & 51.13            \\ 
                                           & SOARISG~\cite{soarisg} $\dagger \ast$  & 12.5 $\pm$ 0.3          & N/A                       & 59.5              & N/A               & 44.6             & N/A              \\ 
                                           & OC-GAN (ours)                                                &$\mathbf{14.0 \pm 0.2}$ & $\mathbf{11.9 \pm 0.5}$            & $\mathbf{36.04}$ & $\mathbf{28.91}$             & $\mathbf{60.32}$            & $\mathbf{58.03}$            \\ \hline
\multirow{1.5}{*}{$128\times128$}          & LostGAN~\cite{isrls} $\dagger$         & 13.8 $\pm$ 0.4          & 11.1 $\pm$ 0.6            & 29.65             & 29.36             & 41.38            & 28.76            \\ 
\multirow{1.5}{*}{with flips}              & LostGAN-V2~\cite{lostganv2} $\dagger$  & 14.2 $\pm$ 0.4         & 10.71 $\pm$ 0.27          & $\mathbf{24.76}$  & 29.00                & 43.27  & 35.17             \\ 
                                           & OC-GAN (ours)                                                & $\mathbf{14.6 \pm 0.4}$ & $\mathbf{12.3 \pm 0.4}$   & 36.31             & $\mathbf{28.26}$  & $\mathbf{59.44}$            & $\mathbf{59.40}$             \\ \hline
\multirow{2}{*}{$256\times256$}            & SOARISG~\cite{soarisg} $\dagger \ast$  & $15.2\pm0.1$           & N/A                       & $65.95$           & N/A               & 45.3              & N/A              \\ 
& OC-GAN (ours) & $\mathbf{17.0 \pm 0.1}$ & $14.4 \pm 0.6$ & $\mathbf{45.96}$ & 39.07 & $\mathbf{53.47}$ & $57.89$ \\ \hline
\multirow{1}{*}{$256\times256$}            & LostGAN-V2~\cite{lostganv2} $\dagger$  & $\mathbf{18.0\pm0.5}$           & $14.1\pm0.4$             & 42.55 & 47.62 &  54.40 & 53.02\\ 
\multirow{1}{*}{with flips}                & 
OC-GAN (ours) & 17.8 $\pm$ 0.2 & $\mathbf{14.7 \pm 0.2}$ & $\mathbf{41.65}$ & $\mathbf{40.85}$ & $\mathbf{57.16}$ & $\mathbf{53.28}$            \\ \hline
\end{tabular}
}
\caption{Performance on $64$, $128$ and $256$ dimension images. All models use ground-truth layouts. We use $\dagger$ to denote results taken from the original paper. $\ast$ denotes a model that uses pixel-level semantic segmentation during training. $\diamond$ denotes models for which the openly available source code was not adapted to generation at a specific image size. We altered the code to allow this and ran a hyperparameter search on the new models. 
}
\label{table:results-table}
\end{table*}

On the proposed SceneFID metric, Table~\ref{table:scene_fid} shows that our method outperforms the others significantly. Thus, our model is significantly better at generating realistic objects compared to the baselines.
Note that the LostGAN model obtains better FID compared to our model exceptionally on $128 \times 128$ COCO-Stuff images but our OC-GAN model outperforms it on the SceneFID metric which is more appropriate in this multi-class setting.
    
\begin{table}
    \resizebox{\columnwidth}{!}{%
    \begin{tabular}{|l|c|c|}\hline
        & \multicolumn{2}{|c|}{SceneFID $\downarrow$} \\\hline
        Methods & COCO & VG \\\hline
        Pix2PixHD~\cite{hd-pix2pix}     & 42.92 & 42.98\\
        SPADE~\cite{spade}              & 23.44 & 16.72\\
        Layout2Im~\cite{layout2im} & 22.76 & 12.56\\
        SOARISG~\cite{soarisg}$\ast$    & 33.46 & N/A\\
        LostGAN~\cite{isrls} (flips)    & 20.03 & 13.17\\
        OC-GAN (ours w/ flips)          & $\mathbf{16.76}$ & $\mathbf{9.63}$\\
        \hline
    \end{tabular} %
    }
    \caption{SceneFID scores on object crops resized to size 224 $\times$ 224, extracted from the 128 $\times$ 128 outputs of the different models, for both datasets. All models use ground-truth layouts. $\ast$ denotes a model that uses pixel-level semantic segmentation during training. SOARISG cannot be trained on VG due to the absence of pixel-level semantic segmentations.
    } 
    \label{table:scene_fid}
\end{table}

\subsection{Qualitative Results}
We compare and analyse image samples generated by our method and competitive baselines in Fig.~\ref{fig:coco-128-compare-models}. In addition to generating higher quality images, our OC-GAN model does not introduce spurious objects (objects not specified in the layout but present in the generated image). This can be attributed to the SGSM module which, by virtue of the retrieval task and the scene-graph being a higher-level abstraction than pixels, aids the model in learning a better mapping from the spatial layout to the generated image. Our model also keeps object instances identifiable even when bounding boxes of objects of the same class overlap slightly or are in close proximity.

To further validate the previous observations, in Fig.~\ref{fig:coco-converging-donuts}, we consider the effect of generating from artificial layouts of gradually converging donuts, to tease out the model's ability to correctly generate separable object instances. Our model generates distinct donuts even when occluded, whereas the other models generate realistic donuts when the bounding boxes are far apart, but fail to do so when they overlap.

We also conducted a user study to evaluate the model's layout-fidelity. 10 users were shown 100 layouts from the test sets of both datasets, with the corresponding images generated by our OC-GAN, LostGAN, and for COCO-Stuff, SOARISG, shuffled in a random order. For each layout, users were asked to select the model which generates the best corresponding image. The results from this study are in Table~\ref{table:user-study} and demonstrate that our model has higher layout-fidelity than previous SOTA methods.

\begin{table}
    \resizebox{\columnwidth}{!}{%
    \begin{tabular}{|l|c|c|c|}\hline
    Dataset & SOARISG & LostGAN & Ours \\
    \hline
    COCO-Stuff & 16.8\% & 36.8\% & \textbf{46.4\%}\\
    VG         & N/R    & 31.4\% & \textbf{68.6\%}\\
    \hline
    \end{tabular} %
    }
    \caption{User study results. 10 computer-science professionals were shown 100 COCO-Stuff and 100 VG test set layouts and corresponding images generated by various models, shuffled randomly. Users were asked to select the highest layout-fidelity image for each layout at $128 \times 128$ resolution. SOARISG is marked marked non-rated (N/R), as it cannot be trained on VG. 
    }
    \label{table:user-study}
\end{table}

\begin{table}[ht]
    \resizebox{\columnwidth}{!}{%
    \begin{tabular}{|l|c|c|}\hline
        & FID $\downarrow$   & CA $\uparrow$  \\ 
        \hline
        \multicolumn{1}{|l|}{Full}                        & $\mathbf{29.57}$ & 60.27  \\ 
        \hline
        \multicolumn{1}{|l|}{Single patchD}               & 30.54 & 59.86  \\
        \multicolumn{1}{|l|}{No patchD}                   & 33.85 & $\mathbf{62.48}$ \\
        \multicolumn{1}{|l|}{No objectD}                  & 31.62 & 48.03  \\ \hline
        \multicolumn{1}{|l|}{No bounding-box instance boundaries}                     & 30.12 & 59.54  \\
        \multicolumn{1}{|l|}{No SGSM}                     & 34.32 & 52.57  \\
        \multicolumn{1}{|l|}{No objectD, no SGSM} & 33.15 & 41.50  \\
        \multicolumn{1}{|l|}{No perceptual loss}          & 31.14 & 57.22  \\
        \multicolumn{1}{|l|}{No perceptual loss, no SGSM} & 36.54 & 47.94  \\
        \hline
    \end{tabular} %
    }
    \caption{Quantitative comparison of different ablated versions of our model on the COCO-Stuff dataset ($64 \times 64$ images). These results highlight the importance of the SGSM (and its positive interaction with the perceptual loss) in the bottom row block, as well as the impact of removing some of the discriminators (middle row block).}
    \label{table:ablation}
\end{table}

In Table~\ref{table:ablation}, we present an ablation study performed by removing certain components of our model. The effect of adding another patch discriminator is measurable, both in terms of FID and CA. Removing the patch discriminator significantly lowers FID (the model has no more supervision in terms of matching the distribution of the real full images. This actually improves the CA, as the generator will use more capacity to focus on generating realistic objects.

We also find that removing either the object discriminator or the SGSM results in a significant drop in performance. This does not however prevent the model from generating realistic objects (the CA score remains above some of the baselines), meaning that the roles of the two components are to some extent complementary. As soon as both are removed, the CA score drops sharply.

Removing the perceptual loss has little effect in itself, but it greatly helps the SGSM when present. Removing the SGSM altogether strongly impairs results, highlighting its importance. Finally, removing the bounding-box instance boundaries has a modest impact on both metrics, but a large qualitative impact with more clearly defined objects.

\section{Conclusion}
We observed that current state-of-the-art layout-to-image generation methods exhibit low layout-fidelity and tend to generate low quality objects especially in cases of occlusion. We proposed a novel Scene-Graph Similarity Module that mitigated the layout-fidelity issues aided by an improved understanding of spatial relationships derived from the layout. We also proposed to condition the generator's normalization layers on instance boundaries which led to sharper, more distinct objects compared to other approaches. The addition of the proposed components to the image generation pipeline led to our model outperforming previous state-of-the-art approaches on a variety of quantitative metrics. A comprehensive ablation study was performed to analyse the contribution of the proposed and existing components of the model. Human users also rated our approach higher on generating better-suited images for the layout over existing methods.

Evaluation metrics for GAN popularized in the single-object-class setting have been criticized as inappropriate in the multi-class setting in literature. Our proposed SceneFID metric addresses those concerns and presents a useful metric for the image generation community which will increasingly deal with multi-class settings in the future. Our proposed OC-GAN model also showed a large improvement over existing approaches on the SceneFID evaluation criteria which further highlights the impact of our contributions.

\section{Acknowledgments}
We acknowledge Emery Fine, Adam Ferguson, Hannes Schulz for their insightful suggestions and valuable assistance. We also thank the many researchers who contributed to the human evaluation study. Finally, we would like to thank the reviewers for their comments and suggestions that helped us improve this manuscript.

{
    \fontsize{9.8pt}{10.8pt}  
    \selectfont
    \bibliography{oc-gan}
}

\newpage

\appendix
\section{Comparison with Semi-Parametric Methods}
Recently, semi-parametric methods have been proposed in the field of layout-to-image generation~\cite{li2019pastegan}. We excluded a comparison with these methods in the main paper due to the fact that (1) they are structurally different (they incorporate real images when generating images) leading to difficulties in making a fair comparison and (2) they function in diverse ways, not all of which can be applied to our setting~\cite{qi2018semi}.

We include a comparison with the state-of-the art semi-parametric model, PasteGAN~\cite{li2019pastegan} in Table~\ref{table:semi-parametric}. This method outperforms most of the other baselines, but still performs worse than our method.

\begin{table}[h]
    \resizebox{\columnwidth}{!}{%
    \begin{tabular}{|l|c|c|}\hline
                                                 & \multicolumn{2}{c|}{Inception Score $\uparrow$} \\\hline
        Methods                                  &                    COCO &                     VG \\\hline
        PasteGAN~\cite{li2019pastegan} $\dagger$ &          $10.2 \pm 0.2$ &          $8.2 \pm 0.2$ \\ 
        OC-GAN (ours)                            & $\mathbf{10.5 \pm 0.3}$ & $\mathbf{8.9 \pm 0.3}$ \\\hline

                                                 & \multicolumn{2}{c|}{FID $\downarrow$} \\\hline
        Methods                                  &             COCO &               VG \\\hline
        PasteGAN~\cite{li2019pastegan} $\dagger$ &          $38.29$ &          $35.25$ \\ 
        OC-GAN (ours)                            & $\mathbf{33.10}$ & $\mathbf{22.61}$ \\\hline
    \end{tabular}
    }
    \caption{Comparison of our method with the semi-parametric method PasteGAN~\cite{li2019pastegan}. We use $\dagger$ to denote results taken from the original paper. The best results in each category are in bold. Our method outperforms this baseline across the evaluation metrics considered.}
    \label{table:semi-parametric}
\end{table}

\section{Dataset statistics}
The dataset statistics are presented in Table~\ref{tb:datasets}.
\begin{table}[h!]
    \resizebox{\columnwidth}{!}{%
    \begin{tabular}{|l|r|r|}\hline
        Dataset & COCO-Stuff & VG\\ \hline
        \# Train Images & 24\,972 & 62\,565\\
        \# Valid Images & 1\,024 & 5\,506\\
        \# Test Images & 2\,048 & 5\,088\\
        \# Objects & 171 & 178\\
        \# Objects in Image & \quad 3 $\sim$ 8 & \quad 3 $\sim$ 30\\
        \hline
    \end{tabular}%
    }
    \caption{Statistics of the COCO-Stuff and Visual Genome datasets.} 
    \label{tb:datasets}
\end{table}

\section{Spatial Relationships used for Generating the Scene-Graph}
We used 6 spatial relationships to generate the scene-graphs from layouts. All of the spatial relationships are derived from the bounding box coordinates specified in the layouts. If an edge in the scene-graph is represented as $<$subject, relationship, object$>$, then the possible relationships we consider are:
\begin{itemize}
    \item ``left of'': subject's centre is to the left of object's centre
    \item ``right of'': subject's centre is to the right of object's centre
    \item ``above'': subject's centre is above object's centre
    \item ``below'': subject's centre is below object's centre 
    \item ``inside'': subject contained inside object
    \item ``surrounding'': object contained inside subject
\end{itemize}

\section{A Note on Evaluation}
Inception Score and FID were computed using the official Tensorflow implementations~\footnote{\url{https://github.com/openai/improved-gan} for Inception Score}\textsuperscript{,}\footnote{\url{https://github.com/bioinf-jku/TTUR} for FID} (the most commonly available PyTorch implementations give slightly different but close values), to ensure compliance with the literature. In the past, papers considering layout and scene graph to image generation have used different values for the number of splits when computing the Inception score, ranging usually from 3 to 5 (as shown in the different official implementations and via contacting some of the authors). Empirically, we found that lowering the split size results in better numerical values for the inception score, for all methods relevant to this work. Out of fairness considerations, we opted for splits of size 5 and note that in addition to this issue, the size of the evaluation set for Inception score computation is very low compared to recommended sizes. This impacts the relevance of this metric.

In addition to the above concerns, some models used different network architectures to compute the inception score (e.g. ~\cite{layout2im} uses a VGG net as opposed to the standard Inception-V3 network as noted in their paper). We used the official Inception-V3-based evaluation on all models.

Some models introduce non-standard data-augmentation (e.g.~\cite{isrls} uses image flips during training). Out of fairness considerations, we compared our approach to the official reported values, and used the same data-augmentation as the compared methods, when applicable.

\section{Complexity of scenes}
We focus on generating images of complex scenes, which warrants a definition of what complex scenes are specifically. In this work, we use the following heuristic. Complex scenes are first and foremost defined with respect to single object datasets: for the most part, images in MS-COCO and VG contain multiple objects (up to 30 in our case). In addition to this, images in both datasets come annotated with relations and attributes (which we do not use in this work, in accordance with the literature). The underlying variability of the scene graphs is also a source of complexity.

\section{Implementation and Training Details}
\begin{figure*}%
    \centering
    \begin{minipage}[b]{.4\linewidth}
        \centering
        {(a) Scene-Graph Encoder \\[1em]\includegraphics[scale=1.0]{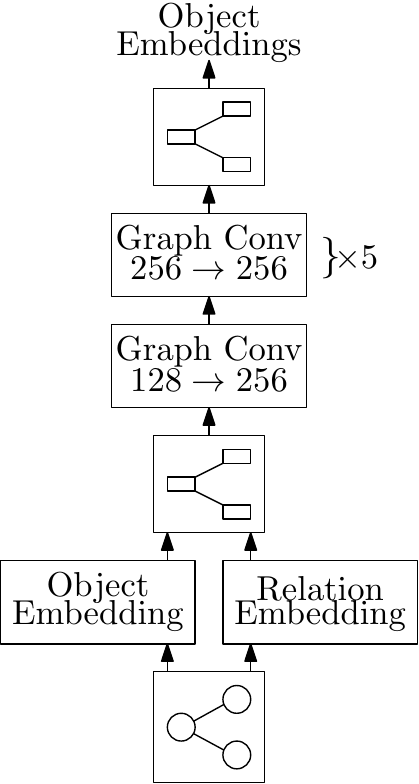}}\\[2em]
        {(b) Conditioning Module \\[1em] \includegraphics[scale=1.0]{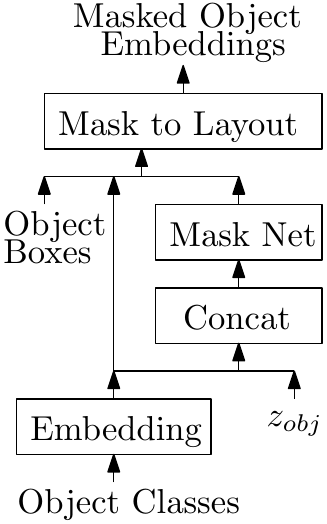}}%
    \end{minipage}%
    \hfill
    \begin{minipage}[b]{.27\linewidth}
        \centering
        {(c) Mask Net \\[1em] \includegraphics[scale=1.0]{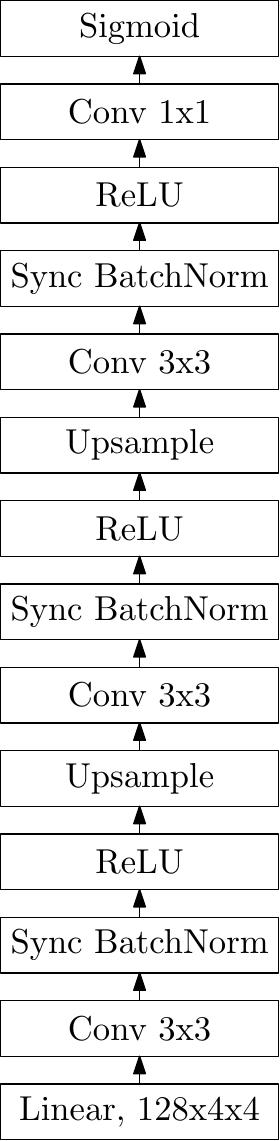}}%
    \end{minipage}%
    \hfill
    \begin{minipage}[b]{.33\linewidth}
        \centering
        {(d) Object Discriminator \\[1em] \includegraphics[scale=1.0]{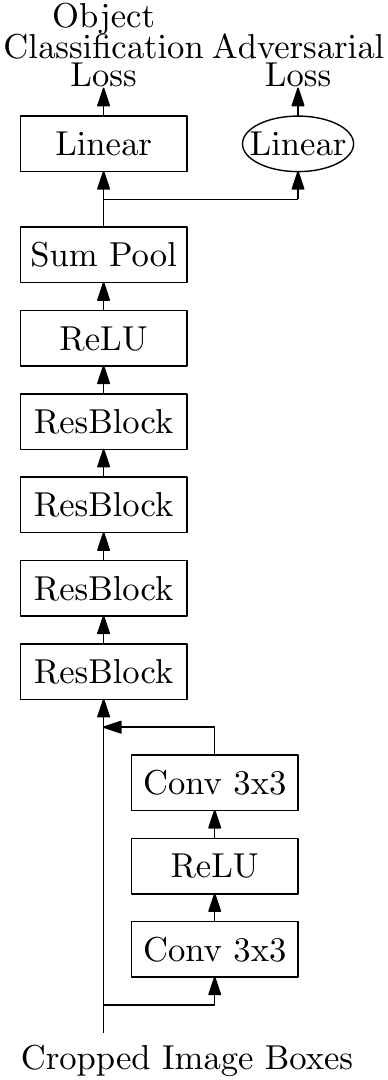}}%
    \end{minipage}%
    \caption{Architecture diagrams for (a) Scene-Graph Encoder (b) Conditioning Module (c)~Mask Net (d) Object Discriminator. The Scene-Graph Encoder takes as input a scene-graph derived from the layout and processes it with a Graph Convolutional Network. The Conditioning Module generates the Masked Object Embeddings, which along with instance boundaries and 1-hot layout, are the conditioning information for the Generator. The Mask Net is a submodule of the Conditioning Module. The Object Discriminator operates on cropped image boxes in an AC-GAN framework, predicting whether the crop is real or generated as well as classifying the object inside the crop.}
    \label{fig:arch-all-others}
\end{figure*}

Architecture diagrams for all the modules of our model OC-GAN are presented in Figs.~\ref{fig:g,g-res,img-d}~and~\ref{fig:arch-all-others}. Some additional hyper-parameter details:
\begin{itemize}
    \item In the SGSM module, images are resized to size $299\times299$ before being processed by the image encoder.
    \item In the SGSM module, the common semantic space for graph and image embeddings has a dimension of $256$.
\end{itemize}

\begin{figure*}%
    \centering
    \begin{minipage}[b]{.35\linewidth}
        \centering
        {(a) Generator \\[1em]\includegraphics[scale=0.9]{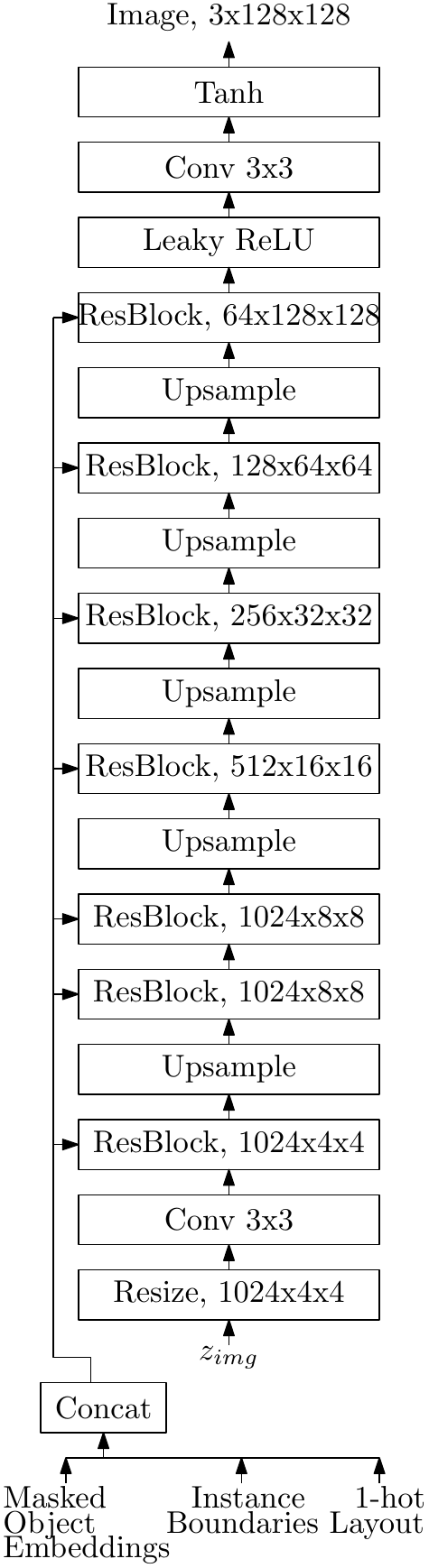}}%
    \end{minipage}%
    \hfill
    \begin{minipage}[b]{.55\linewidth}
        \centering
        {(b) Generator ResBlock \\[1em]\includegraphics[scale=0.9]{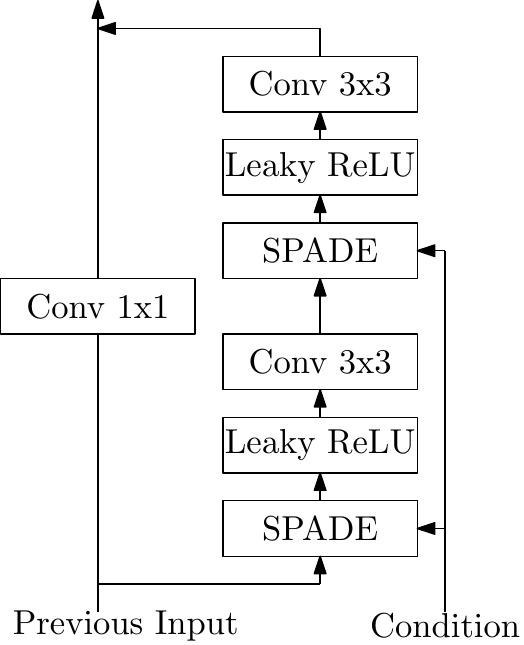}}\\[2em]
        {(c) Image Discriminator \\[1em]\includegraphics[scale=0.9]{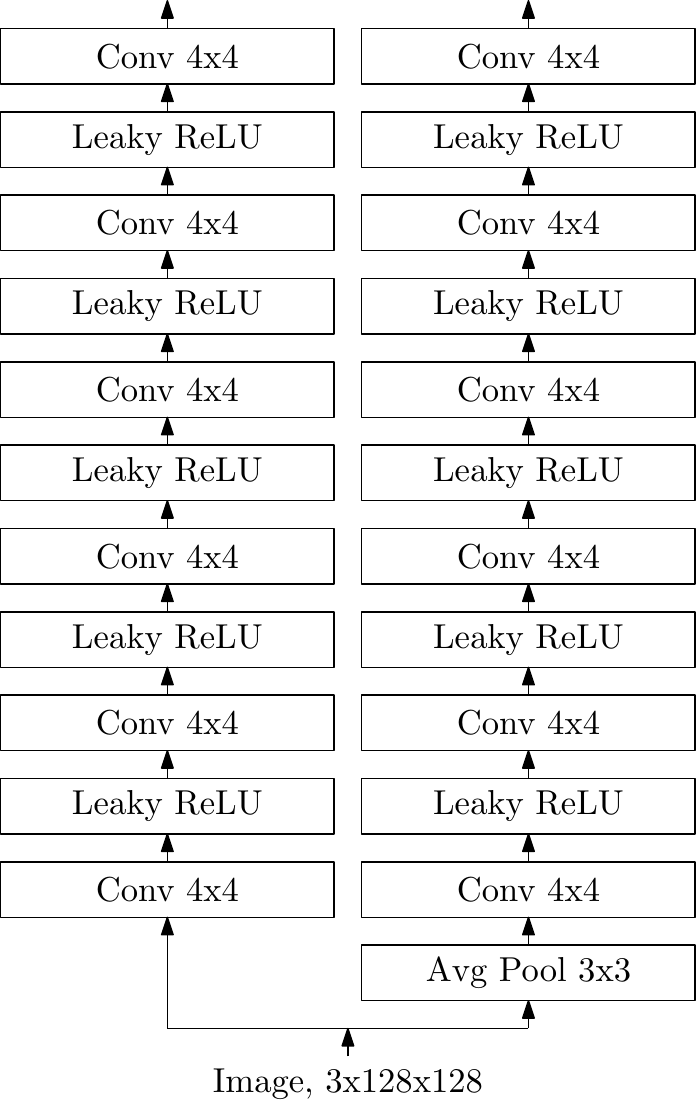}}%
    \end{minipage}%
    \caption{Architecture diagrams for (a) Generator (b) Generator ResBlocks (c) Image Discriminator. All generator inputs are derived from the layout. The Masked Object Embeddings are produced by the Conditioning Module. If input and output dimensions match for the Generator ResBlock, then the shortcut is a skip connection.}%
    \label{fig:g,g-res,img-d}%
\end{figure*}%

\section{Additional Qualitative Results}
\begin{figure}[ht]
    \newcommand{\figwidth}{0.195\columnwidth}
    \newcommand{\figtwidth}{0.19\columnwidth}
    \begin{center}
        \hfill \parbox{\figtwidth}{\centering Layout} \hfill \parbox{\figtwidth}{\centering SPADE} \hfill \parbox{\figtwidth}{\centering SOARISG} \hfill \parbox{\figtwidth}{\centering LostGAN} \hfill \parbox{\figtwidth}{\centering OC-GAN} \hfill\\
        
        \includegraphics[width=\figwidth]{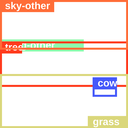}\hfill
        \includegraphics[width=\figwidth]{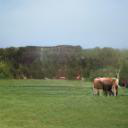}\hfill
        \includegraphics[width=\figwidth]{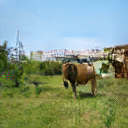}\hfill
        \includegraphics[width=\figwidth]{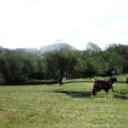}\hfill
        \includegraphics[width=\figwidth]{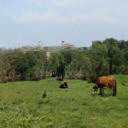}\\
        
        \includegraphics[width=\figwidth]{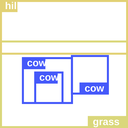}\hfill
        \includegraphics[width=\figwidth]{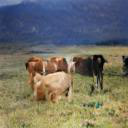}\hfill
        \includegraphics[width=\figwidth]{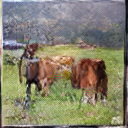}\hfill
        \includegraphics[width=\figwidth]{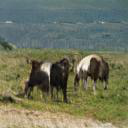}\hfill
        \includegraphics[width=\figwidth]{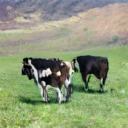}\\
        
        \includegraphics[width=\figwidth]{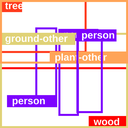}\hfill
        \includegraphics[width=\figwidth]{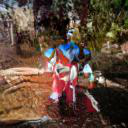}\hfill
        \includegraphics[width=\figwidth]{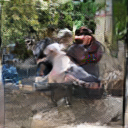}\hfill
        \includegraphics[width=\figwidth]{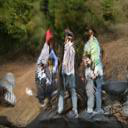}\hfill
        \includegraphics[width=\figwidth]{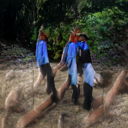}\\
        
        \includegraphics[width=\figwidth]{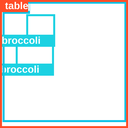}\hfill
        \includegraphics[width=\figwidth]{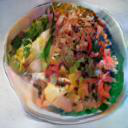}\hfill
        \includegraphics[width=\figwidth]{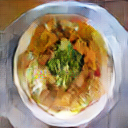}\hfill
        \includegraphics[width=\figwidth]{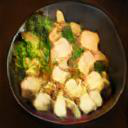}\hfill
        \includegraphics[width=\figwidth]{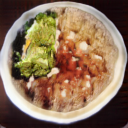}\\
        
        \includegraphics[width=\figwidth]{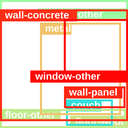}\hfill
        \includegraphics[width=\figwidth]{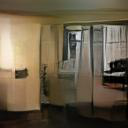}\hfill
        \includegraphics[width=\figwidth]{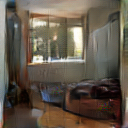}\hfill
        \includegraphics[width=\figwidth]{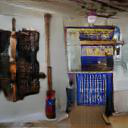}\hfill
        \includegraphics[width=\figwidth]{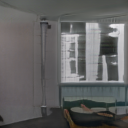}\\
        
        \includegraphics[width=\figwidth]{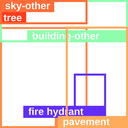}\hfill
        \includegraphics[width=\figwidth]{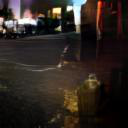}\hfill
        \includegraphics[width=\figwidth]{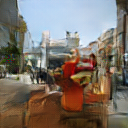}\hfill
        \includegraphics[width=\figwidth]{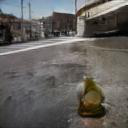}\hfill
        \includegraphics[width=\figwidth]{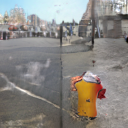}\\
        
        \includegraphics[width=\figwidth]{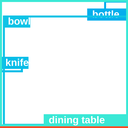}\hfill
        \includegraphics[width=\figwidth]{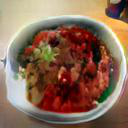}\hfill
        \includegraphics[width=\figwidth]{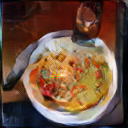}\hfill
        \includegraphics[width=\figwidth]{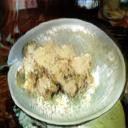}\hfill
        \includegraphics[width=\figwidth]{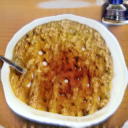}\\
    \end{center}
    \caption{$128 \times 128$ COCO-Stuff test set images, taken from our method (OC-GAN) and multiple competitive baselines.}
    \label{fig:coco-appendix}
\end{figure}

\begin{figure}[!ht]
    \centering
    \newcommand{\figwidth}{0.18\columnwidth}
    \newcommand{\figtwidth}{0.18\columnwidth}
    \begin{center}
        \hfill \parbox{\figtwidth}{\centering Layout} \hfill \parbox{\figtwidth}{\centering LostGAN} \hfill \parbox{\figtwidth}{\centering OC-GAN} \hfill \hspace*{5em}\\
        \hfill
        \includegraphics[valign=m,width=\figwidth]{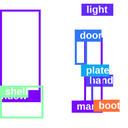}\hfill
        \includegraphics[valign=m,width=\figwidth]{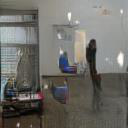}\hfill
        \includegraphics[valign=m,width=\figwidth]{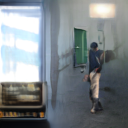}\hfill\hspace*{5em}\\
        \hfill
        \includegraphics[valign=m,width=\figwidth]{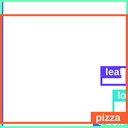}\hfill
        \includegraphics[valign=m,width=\figwidth]{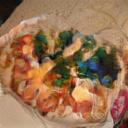}\hfill
        \includegraphics[valign=m,width=\figwidth]{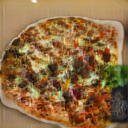}\hfill\hspace*{5em}\\
        \hfill
        \includegraphics[valign=m,width=\figwidth]{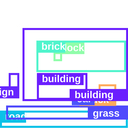}\hfill
        \includegraphics[valign=m,width=\figwidth]{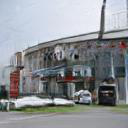}\hfill
        \includegraphics[valign=m,width=\figwidth]{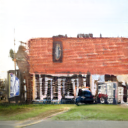}\hfill\hspace*{5em}\\
        \hfill
        \includegraphics[valign=m,width=\figwidth]{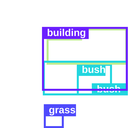}\hfill
        \includegraphics[valign=m,width=\figwidth]{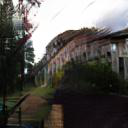}\hfill
        \includegraphics[valign=m,width=\figwidth]{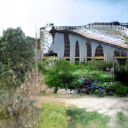}\hfill\hspace*{5em}\\
        \hfill
        \includegraphics[valign=m,width=\figwidth]{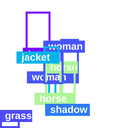}\hfill
        \includegraphics[valign=m,width=\figwidth]{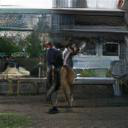}\hfill
        \includegraphics[valign=m,width=\figwidth]{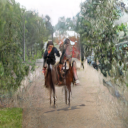}\hfill\hspace*{5em}\\
        \hfill
        \includegraphics[valign=m,width=\figwidth]{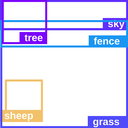}\hfill
        \includegraphics[valign=m,width=\figwidth]{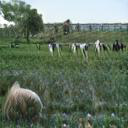}\hfill
        \includegraphics[valign=m,width=\figwidth]{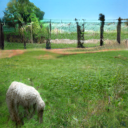}\hfill\hspace*{5em}\\
        \hfill
        \includegraphics[valign=m,width=\figwidth]{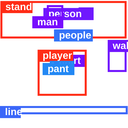}\hfill
        \includegraphics[valign=m,width=\figwidth]{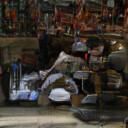}\hfill
        \includegraphics[valign=m,width=\figwidth]{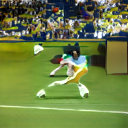}\hfill\hspace*{5em}\\
    \end{center}
    \caption{$128 \times 128$ Visual Genome test set images, taken from our method (OC-GAN) and the LostGAN baseline.}
    \label{fig:vg-appendix}
\end{figure}

We present additional qualitative 128$\times$128 samples on the COCO-Stuff dataset in Fig.~\ref{fig:coco-appendix} and on the Visual Genome dataset in Fig.~\ref{fig:vg-appendix}.

\end{document}